\documentclass[11pt]{article}

\usepackage{acl}

\usepackage{times}
\usepackage{latexsym}
\usepackage[T1]{fontenc}
\usepackage[utf8]{inputenc}
\usepackage{microtype}
\usepackage{inconsolata}
\usepackage{amsmath}
\usepackage{booktabs}
\usepackage{graphicx}
\usepackage{arabtex}

\newcommand{\arpayload}[1]{\mbox{\RL{#1}}}

\makeatletter
\def\endabstract{\unskip\end{list}}
\makeatother

\title{STAGEET: Stage-wise Typed Edit Tagging for Grammatical Error Correction with Arabic as a Case Study}

\author{
Wenjie Lou \quad
Alaa Mamdouh Akef\thanks{Corresponding author: \texttt{alaa\_eldin\_akef@pku.edu.cn}.} \\
School of Foreign Languages, Peking University \\
\texttt{leolouwenjie@gmail.com}
}

\begin{document}
\maketitle

\begin{abstract}
Sequence-to-edit approaches make grammatical error correction (GEC) efficient and locally interpretable by predicting edit labels over the input rather than generating a full corrected sentence. Their interpretability, however, is primarily operational: a label specifies how the string should change, but a single edit vocabulary does not always reveal the type of correction being made. We propose \textsc{STAGEET}, a stage-wise typed edit-tagging framework that reorganizes Seq2Edit supervision into typed executable stages and extends edit operations to correction categories. \textsc{STAGEET} decomposes correction into an ordered sequence of medium-grained typed stages; each stage predicts from its own label space, rewrites the current hypothesis once, and passes the resulting intermediate sentence to the next stage. We instantiate the framework as both an end-to-end shared-encoder multi-head model with stage-specific adapters and a fully specialized variant with one independent tagger per stage. Experiments on QALB-2014 and ZAEBUC show that category-aware staged correction retains competitive edit-based GEC performance while exposing a more inspectable correction trajectory, and attains state-of-the-art results on QALB-2014.
\end{abstract}

\begin{figure*}[t!]
\centering
\includegraphics[width=\textwidth]{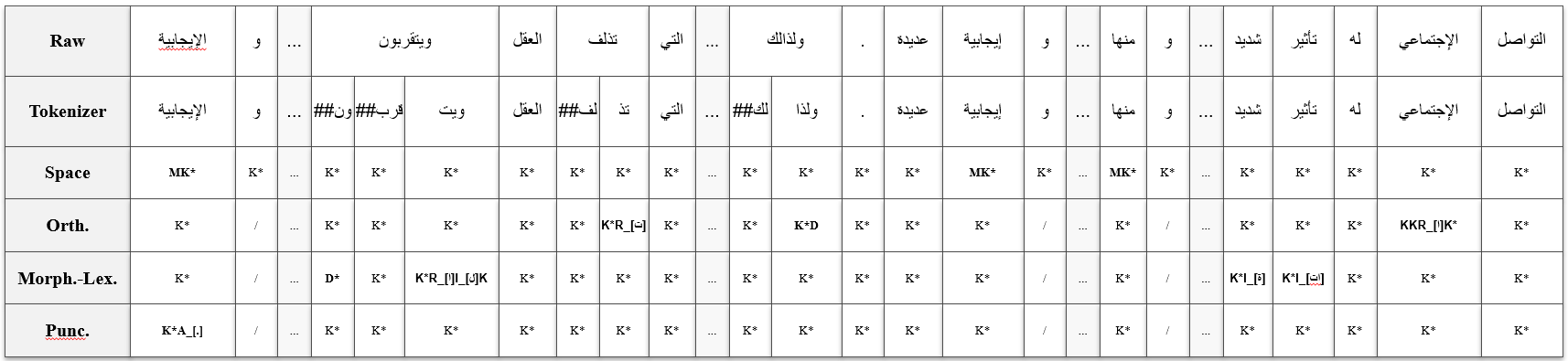}
\caption{An example from \textsc{STAGEET}. Rows show the raw sentence, tokenization, and stage-specific edit labels.}
\label{fig:stageet-example}
\end{figure*}

\section{Introduction}

Grammatical Error Correction (GEC) aims to transform erroneous text into its corrected form. Although the task is often modeled with sequence-to-sequence architectures, it differs from open-ended generation in a consequential respect: most input tokens are copied unchanged, and true corrections are sparse. This property makes text editing an attractive alternative, in which a model predicts token-level edit labels and obtains the corrected sentence by applying them \citep{omelianchuk-etal-2020-gector,alhafni-habash-2025-enhancing}. Such systems are efficient and expose more local decision structure than fully autoregressive generation.

Locality, however, does not by itself yield category-level interpretability. Existing Seq2Edit systems typically place all correction phenomena in a single edit vocabulary. A token may still receive a valid rewrite label even when the underlying correction involves multiple operations or linguistically different phenomena. In that setting, the label explains how to modify the string, but not necessarily what type of error is being corrected.

Arabic makes this limitation especially visible. Corrections may involve clitic boundaries, orthographic variation, rich morphology, lexical choice, and punctuation, often within the same short span. These phenomena are difficult to represent cleanly in one compressed tag inventory.

\textsc{STAGEET} addresses this mismatch by factorizing correction into typed stages. Rather than assigning all edits from a single vocabulary, it constructs stage-specific label spaces and applies them sequentially. The framework uses four stages: spacing, orthography, morpho-lexical correction, and punctuation. Each stage defines its own supervision and produces an intermediate sentence, which makes the correction trajectory inspectable and supports stage-level diagnosis; Figure~\ref{fig:stageet-example} gives a compact token-aligned illustration.

Our contributions are as follows:
\begin{enumerate}
    \item We formulate typed stage-wise edit tagging, a category-aware extension of Seq2Edit that reorganizes edit supervision into typed executable stages with stage-specific label spaces and intermediate hypotheses.
    \item We introduce an automatic construction procedure that derives typed stage supervision from standard raw--corrected GEC sentence pairs, avoiding the need for separately annotated error-type labels.
    \item We instantiate the formulation with a shared-encoder multi-head model and a fully specialized stage-separated variant, and analyze how the induced stages differ in label distribution, cumulative correction behavior, and optimization dynamics.
\end{enumerate}

\section{Background and Related Work}

\subsection{Grammatical Error Correction}

GEC has been approached with rule-based systems, statistical classifiers, statistical machine translation, neural sequence-to-sequence models, edit-based models, and large language models \citep{bryant-etal-2023-grammatical}. English shared tasks and benchmarks such as HOO, CoNLL, JFLEG, and BEA helped move the field toward neural modeling \citep{dale-etal-2012-hoo,ng-etal-2013-conll,ng-etal-2014-conll,napoles-etal-2017-jfleg,bryant-etal-2019-bea}, with sequence-to-sequence GEC later strengthened by synthetic data, copy-augmented architectures, and pretrained encoder-decoder models \citep{yuan-briscoe-2016-grammatical,junczys-dowmunt-grundkiewicz-2016-phrase,junczys-dowmunt-etal-2018-approaching,grundkiewicz-etal-2019-neural,kiyono-etal-2019-empirical,zhao-etal-2019-improving,kaneko-etal-2020-encoder,katsumata-komachi-2020-stronger,rothe-etal-2021-simple}.

GEC is nevertheless structurally different from open-ended generation. The input and output usually have high lexical overlap, and only a small subset of tokens require modification. Full autoregressive decoding can therefore be inefficient and can obscure the local edits responsible for a correction. Evaluation practice also emphasizes edit precision, commonly through F$_{0.5}$ and related edit-based metrics, because overcorrection can be more harmful than leaving some errors unchanged \citep{dahlmeier-ng-2012-better,napoles-etal-2015-ground}.

\subsection{Text Editing for GEC}

Text editing casts GEC as sequence tagging or local transduction: rather than generating the corrected sentence token by token, a model predicts operations that transform the input into the output. Related approaches include LaserTagger \citep{malmi-etal-2019-encode}, PIE \citep{awasthi-etal-2019-parallel}, Seq2Edits \citep{stahlberg-kumar-2020-seq2edits}, and tagging-plus-insertion or instruction-tuned editing models \citep{mallinson-etal-2020-felix,mallinson-etal-2022-edit5,raheja-etal-2023-coedit,raheja-etal-2024-medit}. GECToR showed that a Transformer encoder with token-level edit labels can achieve strong English GEC performance with substantially faster inference than sequence-to-sequence systems \citep{omelianchuk-etal-2020-gector}.

The edit vocabulary is a central design choice. Larger vocabularies express more precise rewrites but increase sparsity and model complexity; smaller vocabularies are easier to learn but may require iterative decoding, hand-designed transformations, or reduced expressiveness. Prior work has therefore explored extended sequence-tagging vocabularies, character-transformation labels, and non-autoregressive editing mechanisms \citep{straka-etal-2021-character,mesham-etal-2023-extended,zhang-etal-2023-non}. In Arabic GEC, SWEET shows that data-derived edit labels can be effective when combined with subword-level representations, edit compression, and pruning \citep{alhafni-habash-2025-enhancing}.

Most text editing systems, however, still use a single monolithic label space for all correction phenomena. A label can therefore be useful as a rewrite instruction without being fully clear as an error category. \textsc{STAGEET} treats this as a label-organization problem: instead of learning one edit vocabulary for all phenomena, it constructs typed, stage-specific vocabularies and exposes intermediate correction states.

\subsection{Arabic Grammatical Error Correction}

Arabic GEC has developed more slowly than English GEC, partly because annotated resources remain limited and partly because Arabic presents rich morphology, clitic attachment, orthographic ambiguity, dialectal variation, and punctuation conventions. These conditions make Arabic a useful case study for typed edit modeling, where category-aware stage supervision must be recovered from ordinary raw--corrected sentence pairs rather than separate manual error-type annotation. The QALB project provided major Modern Standard Arabic correction benchmarks and annotation guidelines \citep{zaghouani-etal-2014-large,mohit-etal-2014-first,zaghouani-etal-2015-correction,rozovskaya-etal-2015-second}; ZAEBUC later introduced a complementary learner-writing domain \citep{habash-palfreyman-2022-zaebuc}.

Earlier Arabic GEC systems used feature-based classifiers, spelling-correction modules, finite-state resources, character- and word-level statistical machine translation, and cascaded systems \citep{eskander-etal-2013-processing,farra-etal-2014-generalized,jeblee-etal-2014-cmuq,rozovskaya-etal-2014-columbia,bougares-bouamor-2015-ummu,nawar-2015-cufe}. More recent work has adopted pretrained sequence-to-sequence models and shown that Arabic GEC benefits from contextual morphological preprocessing and grammatical error detection signals \citep{obeid-etal-2020-camel,inoue-etal-2021-interplay,abdul-mageed-etal-2021-arbert,obeid-etal-2022-camelira,alhafni-etal-2023-advancements}. Prompted large language models have also been studied for GEC, including Arabic GEC and English learner-text correction, with mixed results relative to fine-tuned systems \citep{kwon-etal-2023-beyond,coyne-etal-2023-analyzing,fang-etal-2023-chatgpt,wu-etal-2023-chatgpt,davis-etal-2024-prompting,loem-etal-2023-exploring}.

Text editing for Arabic has only recently become competitive. SWEET provides a strong monolithic edit-tagging baseline and shows that subword-level compressed edits with pruning offer a practical balance between coverage and learnability \citep{alhafni-habash-2025-enhancing}. The present work asks a different question: how the edit space itself should be organized so that correction becomes a sequence of typed executable stages rather than one undifferentiated inventory.

\subsection{Edit Tagging and Interpretability}

Interpretability in GEC is often addressed after correction, by annotating edits with error types and then analyzing system behavior. Linguistically informed edit extraction supports reliable learner-error analysis \citep{felice-etal-2016-automatic}; ERRANT enables detailed English error-type evaluation beyond aggregate M2 scores \citep{bryant-etal-2017-automatic}; and ARETA provides Arabic error-type annotation designed around morphological richness and orthographic ambiguity \citep{belkebir-habash-2021-automatic}. Error detection work similarly shows that category granularity affects modeling and interpretation \citep{yuan-etal-2021-multi}.

Post-hoc error annotation and edit-tagging models occupy different parts of the modeling stack. In standard edit tagging, the model predicts rewrite commands, and linguistic categories are recovered only afterward, if at all. Token-level locality is therefore distinct from category-level interpretability: a local edit label can still mix several linguistic phenomena, especially when labels are compressed or when a single token participates in multiple operations. Unlike prior work that uses error types mainly for analysis, evaluation, prompting, or system selection, \textsc{STAGEET} makes correction categories part of Seq2Edit supervision and decoding through typed executable stages.

\section{Approach}

\textsc{STAGEET} is a typed text-editing framework. Given an erroneous sentence $x$ and a corrected sentence $y$, the objective is not to learn a single edit sequence that maps $x$ directly to $y$. Instead, correction is decomposed into a sequence of typed intermediate targets:
\[
x_0 \rightarrow x_1 \rightarrow x_2 \rightarrow x_3 \rightarrow x_4,
\]
where $x_0$ is the raw sentence and $x_4$ is the final corrected sentence. Each transition corresponds to a distinct correction stage.

A central design choice is that the type system is part of the executable correction process rather than a post-hoc analysis layer. Each stage has its own supervision, label vocabulary, prediction component, and intermediate output. Training therefore targets not only the rewrite itself, but also the typed correction layer responsible for that rewrite.

\subsection{Typed Edit Construction}

Standard GEC datasets provide raw and corrected sentences, but not stage-wise error-type supervision. \textsc{STAGEET} begins with an automatic construction step: raw--gold sentence pairs are aligned and transformed into edit operations, and each operation is assigned to a typed stage according to its linguistic function. The result is a sequence of stage targets rather than a single monolithic target.

This construction allows ordinary parallel GEC data to support typed correction without manual error-type annotation.

\subsection{Stage Taxonomy}

The stage taxonomy is deliberately medium-grained. Very coarse categories would provide limited interpretive value, whereas very fine-grained categories would create sparse label spaces. The current taxonomy contains four stages.

\paragraph{Spacing}
The spacing stage handles word-boundary and clitic-boundary edits, including merges and splits.

\paragraph{Orthography}
The orthography stage covers local spelling and orthographic variants, including hamza, alif forms, ya/alif maqsura, ta marbuta, and common character-level errors.

\paragraph{Morpho-lexical correction}
The morpho-lexical stage captures broader residual grammatical edits, including inflectional changes, agreement-related changes, lexical substitutions, and local rewrites that are not purely orthographic. In Arabic, morphology and lexical choice are often tightly coupled through root-and-pattern derivation and cliticized forms, so many corrections cannot be cleanly assigned to morphology or lexical substitution alone.

\paragraph{Punctuation}
The punctuation stage handles insertion, deletion, and replacement of punctuation marks. It is placed last because punctuation decisions differ from lexical and orthographic edits in both distribution and certainty. Such edits are sparse, highly precision-sensitive, and partly shaped by annotator preference: in many contexts, more than one punctuation choice may be acceptable even when the reference records only one.

\subsection{Stage Order and Decoding}

Stages are applied in a fixed order. Spacing comes first because token boundaries affect later alignment and edit projection. Orthography follows, capturing local character-level variation. Morpho-lexical correction is applied after these surface-normalizing stages because it may involve broader word-level choices. Punctuation is applied last so that it does not interfere with lexical correction and can be treated as a final, annotation-sensitive decision point. Before this final stage, the intermediate hypothesis is retokenized with punctuation marks separated from neighboring tokens, which prevents punctuation edits from being absorbed into lexical or subword edits during decoding.

At stage $k$, the model observes $x_{k-1}$ and predicts edit labels $e^k$:
\[
e^k = f_k(x_{k-1}).
\]
Applying these edits yields the next intermediate sentence:
\[
x_k = \mathrm{Apply}(x_{k-1}, e^k).
\]
Each stage is applied exactly once. \textsc{STAGEET} therefore performs non-iterative staged decoding rather than repeatedly applying the same tagger until convergence.

The sequential design also avoids an ambiguity that would arise if a single tagger predicted all typed edit categories at once. When multiple typed edits attach to the same token or span, joint application can create reconstruction conflicts. The problem is especially acute when one edit type merges, splits, or deletes a token: other typed edits attached to that token may lose their anchor or refer to a different surface form after reconstruction. By applying one typed edit space at a time, \textsc{STAGEET} re-anchors later predictions on the intermediate sentence and produces a well-defined final hypothesis for M2 scoring.

\subsection{Edit Representation}

Each stage has its own label vocabulary, preserving stage-level label interpretation and reducing the burden on any single edit inventory to cover heterogeneous phenomena. Edits are extracted at the subword level and compressed into frequent rewrite patterns, while rare labels can be pruned during training. Unlike monolithic edit tagging, the same surface operation is interpreted within a typed context: an insertion in the spacing stage and an insertion in the punctuation stage correspond to different correction decisions.

\subsection{Model Variants}

\paragraph{\textsc{STAGEET-MH}}
The main model uses a shared pretrained Transformer encoder with stage-specific adapters and heads. Given input $x_{k-1}$ at stage $k$, the encoder produces contextual representations:
\[
H^k = \mathrm{Encoder}(x_{k-1}).
\]
The stage-specific adapter transforms the shared representation:
\[
\tilde{H}^k = \mathrm{Adapter}_k(H^k),
\]
and the corresponding head predicts a distribution over the stage-specific label set $\mathcal{L}_k$:
\[
p(e_i^k \mid x_{k-1}) =
\mathrm{softmax}(W_k \tilde{h}_i^k + b_k).
\]
The heads do not share output vocabularies. The shared encoder keeps the model compact, while the adapters provide stage-specific capacity.

Training is end-to-end over the shared encoder, stage-specific adapters, and stage-specific heads. The objective sums the token-level losses from the typed stages, so all trainable components in \textsc{STAGEET-MH} are optimized jointly from automatically constructed supervision. Edit application is performed only during staged decoding at inference time.

\paragraph{\mbox{\textsc{STAGEET-Sep}}}
A fully specialized variant trains one independent tagger per stage. Each tagger is trained end-to-end for its own typed objective, while the stages are applied sequentially at inference time. Removing encoder sharing gives each stage its own encoder-level representation, at the cost of a larger parameter footprint.

\section{Experimental Setup}

\subsection{Data}

Experiments are conducted on QALB-2014 and ZAEBUC, two Arabic GEC benchmarks with different domains. QALB-2014 contains Modern Standard Arabic errors from the shared-task setting, whereas ZAEBUC consists of learner essays. For QALB-2014, we use the official shared-task split and M2 annotations \citep{mohit-etal-2014-first}. For ZAEBUC, we use the split and alignment-derived M2 edits released with the Arabic GEC setup of \citet{alhafni-etal-2023-advancements}, since the original ZAEBUC corpus provides raw--corrected sentence pairs rather than manually authored M2 edits \citep{habash-palfreyman-2022-zaebuc}.

\subsection{Evaluation}

Systems are evaluated with the MaxMatch (M2) scorer \citep{dahlmeier-ng-2012-better}, which computes precision, recall, F1, and F$_{0.5}$ over edits. F$_{0.5}$ places greater weight on precision, reflecting the common GEC preference for reliable corrections over aggressive overcorrection. Beyond final M2 scores, the staged formulation enables cumulative evaluation after stage prefixes and separate analysis before the punctuation stage.

\subsection{Baselines}

The comparison covers representative Arabic GEC systems from several modeling families. A'2023 (Seq2Seq) and A'2023 (Seq2Seq++) represent neural generation systems, with the latter incorporating additional Arabic preprocessing and grammatical error detection features \citep{alhafni-etal-2023-advancements}. GPT-4o \citep{openai-2024-gpt4o}, Fanar \citep{fanar-team-2025-fanar}, and Jais-13B-Chat \citep{sengupta-etal-2023-jais} are included as prompted LLM references under Arabic prompting settings rather than primary fine-tuned baselines. SWEET is included on the development sets as a monolithic Seq2Edit reference \citep{alhafni-habash-2025-enhancing}; we do not report SWEET test results because the publicly comparable setting is development-set evaluation, while available test-side SWEET results involve iterative variants that are not fully matched to our single-iteration setting.

\subsection{Implementation Details}

AraBERTv02 \citep{antoun-etal-2020-arabert} serves as the encoder backbone. In \textsc{STAGEET-MH}, the encoder, adapters, and stage-specific heads are optimized jointly with an end-to-end multi-task tagging objective. In \mbox{\textsc{STAGEET-Sep}}, each stage tagger is trained end-to-end but separately. Checkpoints are selected on development data using M2: the multi-head model uses the best development checkpoint, while the sequential variant retains top stage candidates and composes them sequentially. This gives the sequential variant a larger model-selection budget, so it is interpreted as a fully specialized staged variant, while \textsc{STAGEET-MH} is the compact shared-encoder instantiation. Rare edit labels are pruned with stage-specific thresholds. The punctuation confidence threshold is selected on development data from a fixed candidate set; the selected value is 0.6 and is then applied unchanged to test evaluation. 
\section{Results and Analysis}

\subsection{Main Results}

Final M2 precision, recall, F1, and F$_{0.5}$ are reported on QALB-2014 and ZAEBUC. Table~\ref{tab:dev-results} presents development results. \textsc{STAGEET-MH} denotes the shared-encoder multi-head model, while \mbox{\textsc{STAGEET-Sep}} denotes the fully specialized four-tagger variant. Unless otherwise stated, both variants use the development-selected punctuation confidence threshold.

\begin{table*}[t]
\centering
\small
\begin{tabular}{lcccccccc}
\toprule
 & \multicolumn{4}{c}{QALB-2014 Dev} & \multicolumn{4}{c}{ZAEBUC Dev} \\
\cmidrule(lr){2-5}\cmidrule(lr){6-9}
System & P & R & F1 & F$_{0.5}$ & P & R & F1 & F$_{0.5}$ \\
\midrule
A'2023 (Seq2Seq) & 83.2 & 64.9 & 72.9 & 78.7 & 87.3 & 70.6 & 78.1 & 83.4 \\
A'2023 (Seq2Seq++) & 83.1 & 67.9 & \textbf{74.7} & 79.6 & \textbf{87.6} & 73.9 & \textbf{80.2} & \textbf{84.5} \\
\midrule
GPT-4o & 79.7 & 67.3 & 73.0 & 76.9 & 83.4 & \textbf{75.6} & 79.3 & 81.7 \\
Fanar & 69.5 & 64.4 & 66.8 & 68.4 & 75.3 & 74.0 & 74.6 & 75.0 \\
Jais-13B-Chat & 48.2 & 37.5 & 42.2 & 45.6 & 50.1 & 20.9 & 29.5 & 39.2 \\
\midrule
SWEET & 81.8 & \textbf{68.8} & \textbf{74.7} & 78.8 & 85.8 & 72.3 & 78.4 & 82.7 \\
\midrule
\textbf{\textsc{STAGEET-MH}} & 84.3 & 65.0 & 73.4 & 79.6 & 86.7 & 71.1 & 78.1 & 83.0 \\
\textbf{\mbox{\textsc{STAGEET-Sep}}} & \textbf{85.9} & 64.4 & 73.6 & \textbf{80.5} & 86.2 & 71.2 & 78.0 & 82.7 \\
\bottomrule
\end{tabular}
\caption{Development results on QALB-2014 and ZAEBUC. A'2023 denotes \citet{alhafni-etal-2023-advancements}. Best overall result in each metric is in bold.}
\label{tab:dev-results}
\end{table*}

Table~\ref{tab:test-results} reports test results for the two \textsc{STAGEET} variants and the remaining comparison systems.

\begin{table*}[t]
\centering
\small
\begin{tabular}{lcccccccc}
\toprule
 & \multicolumn{4}{c}{QALB-2014 Test} & \multicolumn{4}{c}{ZAEBUC Test} \\
\cmidrule(lr){2-5}\cmidrule(lr){6-9}
System & P & R & F1 & F$_{0.5}$ & P & R & F1 & F$_{0.5}$ \\
\midrule
A'2023 (Seq2Seq) & 84.0 & 64.7 & 73.1 & 79.3 & \textbf{86.0} & 71.6 & 78.2 & 82.7 \\
A'2023 (Seq2Seq++) & 84.2 & 65.4 & \textbf{73.6} & 79.6 & 85.9 & 73.4 & \textbf{79.2} & \textbf{83.1} \\
\midrule
GPT-4o & 80.4 & \textbf{66.8} & 73.0 & 77.3 & 81.7 & \textbf{74.9} & 78.2 & 80.2 \\
\midrule
\textbf{\textsc{STAGEET-MH}} & 84.7 & 63.6 & 72.7 & 79.4 & 85.3 & 71.5 & 77.8 & 82.1 \\
\textbf{\mbox{\textsc{STAGEET-Sep}}} & \textbf{86.0} & 62.8 & 72.6 & \textbf{80.1} & 84.7 & 72.6 & 78.2 & 81.9 \\
\bottomrule
\end{tabular}
\caption{Test results on QALB-2014 and ZAEBUC. A'2023 denotes \citet{alhafni-etal-2023-advancements}. Best overall result in each metric is in bold.}
\label{tab:test-results}
\end{table*}

The results indicate that typed staged correction provides an inspectable correction trajectory while maintaining competitive GEC performance. On QALB-2014, \mbox{\textsc{STAGEET-Sep}} obtains the highest precision-oriented score among the reported systems, showing that the added stage structure can support a conservative correction profile. On ZAEBUC, the results are closer across systems, and the strongest scores remain split across model families, which suggests that the learner-essay domain places different pressure on the precision-recall balance. The two \textsc{STAGEET} variants occupy different points in the staged design space: \textsc{STAGEET-MH} keeps all stages in a compact shared model, whereas \mbox{\textsc{STAGEET-Sep}} gives each stage a fully specialized tagger.

\subsection{Runtime and Model Size}

Table~\ref{tab:runtime} compares model size and inference runtime on QALB-2014 development data. \textsc{STAGEET-MH} stores a single shared encoder with stage-specific adapters and heads, yielding a compact parameter footprint. \mbox{\textsc{STAGEET-Sep}} stores four independent taggers, increasing stored parameters while retaining the same staged decoding depth.

Compared with prompted large language models such as GPT-4o, the \textsc{STAGEET} variants are substantially smaller and faster to run, making them more suitable for deployment-constrained settings.

\begin{table}[t]
\centering
\small
\setlength{\tabcolsep}{4pt}
\begin{tabular}{lrr}
\toprule
System & Params & Runtime \\
\midrule
A'2023 (Seq2Seq) & 139M & 71 \\
A'2023 (Seq2Seq++) & 502M & 219 \\
\midrule
\textsc{STAGEET-MH} & 135M & 46 \\
\mbox{\textsc{STAGEET-Sep}} & 540M & 46 \\
\bottomrule
\end{tabular}
\caption{Number of parameters (Params.) and inference runtime on QALB-2014 development data. Runtime is reported in seconds using a batch size of 32.}
\label{tab:runtime}
\end{table}

\subsection{Stage-wise Label Distributions}

The staged construction reorganizes edit labels into category-aware stage-specific vocabularies without relying on manually annotated error types. We inspect the induced training labels to characterize how correction phenomena are distributed across the different stages.

Table~\ref{tab:label-stats} reports statistics from the constructed training data under two training configurations: QALB-2014 alone and QALB-2014 combined with ZAEBUC. In both settings, non-KEEP labels account for only a small fraction of token positions, confirming the sparsity of correction in edit-based GEC. At the same time, the stages differ substantially in label distribution. Orthography has the largest active-token share, at around 12\%, reflecting frequent local form variation. The residual morpho-lexical stage is more heterogeneous: it produces the largest raw label inventory, but pruning reduces this long tail to a much smaller set of trainable labels. Spacing and punctuation have more compact vocabularies, consistent with their narrower edit functions.

These statistics support the central label-organization claim. A monolithic edit vocabulary must represent these distributions in a single space, so an edit label can be operationally valid while remaining category-mixed. \textsc{STAGEET} instead assigns each label vocabulary to one typed stage. The same surface operation, such as insertion or deletion, is therefore interpreted with respect to its stage, yielding correction labels that are executable and category-aware.

\begin{table*}[t]
\centering
\small
\begin{tabular}{llrrrr}
\toprule
Dataset & Stage & Examples with edits & Non-KEEP tokens & Label types & Kept labels \\
\midrule
QALB-2014 & Spacing & 56.8\% & 2.4\% & 72 & 11 \\
QALB-2014 & Orthography & 93.0\% & 12.3\% & 1339 & 298 \\
QALB-2014 & Morpho-lexical & 58.9\% & 2.5\% & 2621 & 219 \\
QALB-2014 & Punctuation & 96.8\% & 9.8\% & 150 & 32 \\
\midrule
QALB-2014+ZAEBUC & Spacing & 57.0\% & 2.4\% & 75 & 11 \\
QALB-2014+ZAEBUC & Orthography & 93.1\% & 12.3\% & 1378 & 307 \\
QALB-2014+ZAEBUC & Morpho-lexical & 59.2\% & 2.5\% & 2698 & 222 \\
QALB-2014+ZAEBUC & Punctuation & 96.8\% & 9.7\% & 157 & 34 \\
\bottomrule
\end{tabular}
\caption{Stage-wise training-label statistics. Examples with edits is the percentage of training examples containing at least one operation for the stage. Non-KEEP tokens is the percentage of token positions assigned a non-KEEP edit label. Kept labels counts labels retained after frequency pruning, including KEEP, with thresholds 10/10/10/30 for spacing, orthography, morpho-lexical, and punctuation.}
\label{tab:label-stats}
\end{table*}

\subsection{Label Purity}

Because the stage supervision is constructed automatically, we also examine the purity of the induced labels, namely whether non-\texttt{K*} edits correspond to their intended categories. An expert manually inspects the non-\texttt{K*} labels in 50 randomly sampled training sentences, including 40 from QALB-2014 and 10 from ZAEBUC. Table~\ref{tab:manual-stage-assignment} reports the resulting label-assignment quality.

\begin{table}[t]
\centering
\small
\setlength{\tabcolsep}{3pt}
\begin{tabular}{lrrr}
\toprule
Stage & Mismatch & Boundary & Accept. \\
\midrule
Space & 0.0 & 6.8 & 100.0 \\
Punc. & 0.0 & 0.0 & 100.0 \\
Orth. & 1.4 & 0.0 & 98.6 \\
Morph. & 5.1 & 8.6 & 94.9 \\
\bottomrule
\end{tabular}
\caption{Manual inspection of induced stage labels. Values are percentages over non-\texttt{K*} labels. Boundary cases are not treated as unacceptable assignments.}
\label{tab:manual-stage-assignment}
\end{table}

The clear mismatches are concentrated in orthography and morpho-lexical. In orthography, they come from verbal suffix normalization, where a local character insertion such as the Arabic plural suffix marker is better viewed as morpho-lexical. The morpho-lexical mismatches are mostly substantial local rewrites that remain clearly attributable to spelling, such as repeated letters or other overt orthographic errors. Boundary and alignment cases mostly reflect non-unique raw--gold alignments, where one local operation may be anchored to a neighboring token even though the overall stage assignment remains acceptable.

\subsection{Cumulative Stage Behavior}

The staged formulation also permits evaluation after each prefix of the correction trajectory. This is not a pure error-type evaluation, since M2 is computed against the full gold correction and later stages may repair errors that earlier stages intentionally leave untouched. It is nevertheless useful as a cumulative diagnostic: each row shows the quality of the hypothesis after applying a longer prefix of typed stages.

Table~\ref{tab:cumulative-stage} reports QALB-2014 development results before punctuation is applied. The spacing-only stage has high precision but very low recall, as expected: it targets a narrow set of boundary edits and leaves most gold corrections unresolved. Adding orthography produces the largest jump in F$_{0.5}$, raising both variants above 0.73. The morpho-lexical stage then provides a smaller but consistent gain, mainly through improved recall. The same trend appears in both model variants, while the fully specialized variant remains slightly stronger after the orthography and morpho-lexical prefixes.

\begin{table}[t]
\centering
\small
\setlength{\tabcolsep}{3.5pt}
\begin{tabular}{llrrrr}
\toprule
Model & Prefix & P & R & F1 & F$_{0.5}$ \\
\midrule
MH & Spacing & 70.0 & 6.5 & 11.9 & 23.8 \\
MH & +Orth. & 87.7 & 44.8 & 59.3 & 73.6 \\
MH & +Morph. & 86.8 & 47.5 & 61.4 & 74.5 \\
\midrule
Sep & Spacing & 72.5 & 6.4 & 11.7 & 23.6 \\
Sep & +Orth. & 88.9 & 44.7 & 59.5 & 74.2 \\
Sep & +Morph. & 88.0 & 47.5 & 61.7 & 75.2 \\
\bottomrule
\end{tabular}
\caption{Cumulative QALB-2014 development performance after stage prefixes, before punctuation. MH denotes \textsc{STAGEET-MH}; Sep denotes \mbox{\textsc{STAGEET-Sep}}. M2 is computed against the full gold correction, so early prefixes are diagnostic rather than standalone systems.}
\label{tab:cumulative-stage}
\end{table}

\subsection{Gradient Interference}

The multi-head model shares an encoder across stages, but the stages differ in label space, error distribution, and decision boundary. We measure pairwise gradient alignment among stage losses on shared BERT parameters. Let $g_i = \nabla_{\theta}\mathcal{L}_i$ and $g_j = \nabla_{\theta}\mathcal{L}_j$ be gradients from two stages with respect to shared parameters $\theta$. Their cosine similarity is:
\[
\mathrm{cos}(g_i,g_j) =
\frac{g_i \cdot g_j}{\|g_i\|\|g_j\|},
\]
and a pair is treated as conflicting when $\mathrm{cos}(g_i,g_j) < 0$.

\begin{table}[t]
\centering
\small
\begin{tabular}{lcc}
\toprule
Parameter block & Avg. cosine & Conflict rate \\
\midrule
Full BERT & .003--.013 & 20--50\% \\
Last BERT layer & -.004--.001 & 48--56\% \\
\bottomrule
\end{tabular}
\caption{Summary of gradient interference among stage losses. Average cosine reports the range of off-diagonal pairwise cosine similarities in the representative measurement detailed in Appendix~\ref{sec:gradient-conflict-matrices}. Conflict rate is the proportion of stage-pair gradients with negative cosine similarity.}
\label{tab:gradient-conflict}
\end{table}

The measurement indicates that stage gradients are only weakly aligned. Across the full BERT encoder, pairwise cosine similarities are positive but close to zero, with conflict rates ranging from 20\% to 50\%. In the last BERT layer, the cosines become slightly negative for most stage pairs and the conflict rates are consistently higher, ranging from 48\% to 56\%. This pattern is consistent with the view that stage objectives place different pressures on the shared representation near the prediction interface, and it motivates stage-specific adapters as a compact way to provide private capacity while retaining a shared encoder.

Table~\ref{tab:adapter-ablation} further compares the shared multi-head model with and without stage-specific adapters on QALB-2014 development data. The adapter-based variant improves overall performance, suggesting that stage-specific capacity is useful even when the encoder is shared across all stages.

\begin{table}[t]
\centering
\small
\setlength{\tabcolsep}{4pt}
\begin{tabular}{lrrrr}
\toprule
Model & P & R & F1 & F$_{0.5}$ \\
\midrule
Shared-MH (w/o adapters) & 84.1 & 62.3 & 71.6 & 78.6 \\
Shared-MH (w/ adapters) & 84.3 & 65.0 & 73.4 & 79.6 \\
\bottomrule
\end{tabular}
\caption{Adapter ablation on QALB-2014 development data.}
\label{tab:adapter-ablation}
\end{table}

\subsection{Punctuation Confidence}

Punctuation differs qualitatively from many lexical and orthographic errors. In many contexts, punctuation does not have a single universally correct realization: annotators may differ in comma placement, sentence splitting, quotation conventions, or whether punctuation should be inserted at all. Matching gold punctuation is therefore partly shaped by annotation preference. This motivates the use of a fixed confidence threshold for punctuation predictions in the final stage. The staged formulation treats punctuation as a separate final decision point, allowing punctuation edits to be applied only when the model assigns sufficiently high confidence in its edit predictions.

\section{Conclusion and Future Work}

This paper introduced \textsc{STAGEET}, a stage-wise typed edit-tagging framework for grammatical error correction. Rather than placing all rewrite operations in a single monolithic edit vocabulary, \textsc{STAGEET} decomposes correction into ordered typed stages, each with its own label space, prediction component, and intermediate output. The design preserves the efficiency and local editability of Seq2Edit models while making the correction process more category-aware and inspectable.

Experiments on QALB-2014 and ZAEBUC show that staged typed correction is an effective formulation for Arabic GEC. The shared multi-head model provides a compact end-to-end instantiation, while the fully specialized variant illustrates the effect of stronger stage separation. The induced stages differ not only in linguistic scope, but also in label sparsity and optimization behavior. Future work can adapt the taxonomy to other languages, explore reliable single-pass typed tagging, and develop more targeted punctuation modeling.

\section*{Limitations}

The experiments focus on Arabic GEC. Although the stage-wise formulation is general, the current taxonomy is designed around Arabic error phenomena. Applying \textsc{STAGEET} to other languages would require defining appropriate typed stages. In addition, \textsc{STAGEET} introduces multiple decoding steps, although each stage is applied only once. A related alternative would be to predict typed edit labels for all error categories in a single forward pass rather than through staged decoding. This design has potential, but it complicates quantitative evaluation: multiple typed edits may attach to the same token or span, and their joint reconstruction can create conflicts before M2 scoring. We therefore leave single-pass typed tagging as future work.

\section*{Ethical Considerations}

This work uses publicly available or previously released Arabic GEC datasets and evaluates systems on text-correction benchmarks. The proposed method is intended for grammatical error correction and analysis rather than for making high-stakes decisions about writers. Because GEC systems may alter user text in ways that affect meaning or style, corrected outputs should be reviewed in educational or professional settings where accuracy is important. Prompted LLM references are evaluated only as comparison systems; no private user data is submitted to external services as part of the experiments.

\bibliography{custom}

@article{bryant-etal-2023-grammatical,
  title = {Grammatical Error Correction: A Survey of the State of the Art},
  author = {Bryant, Christopher and Yuan, Zheng and Qorib, Muhammad Reza and Cao, Hannan and Ng, Hwee Tou and Briscoe, Ted},
  journal = {Computational Linguistics},
  volume = {49},
  number = {3},
  pages = {643--701},
  year = {2023},
  doi = {10.1162/coli\_a\_00478},
  url = {https://aclanthology.org/2023.cl-3.4/}
}

@inproceedings{ng-etal-2014-conll,
  title = {The {CoNLL}-2014 Shared Task on Grammatical Error Correction},
  author = {Ng, Hwee Tou and Wu, Siew Mei and Briscoe, Ted and Hadiwinoto, Christian and Susanto, Raymond Hendy and Bryant, Christopher},
  booktitle = {Proceedings of the Eighteenth Conference on Computational Natural Language Learning: Shared Task},
  pages = {1--14},
  year = {2014},
  address = {Baltimore, Maryland},
  publisher = {Association for Computational Linguistics},
  doi = {10.3115/v1/W14-1701},
  url = {https://aclanthology.org/W14-1701/}
}

@inproceedings{bryant-etal-2019-bea,
  title = {The {BEA}-2019 Shared Task on Grammatical Error Correction},
  author = {Bryant, Christopher and Felice, Mariano and Andersen, {\O}istein E. and Briscoe, Ted},
  booktitle = {Proceedings of the Fourteenth Workshop on Innovative Use of NLP for Building Educational Applications},
  pages = {52--75},
  year = {2019},
  address = {Florence, Italy},
  publisher = {Association for Computational Linguistics},
  doi = {10.18653/v1/W19-4406},
  url = {https://aclanthology.org/W19-4406/}
}

@inproceedings{dahlmeier-ng-2012-better,
  title = {Better Evaluation for Grammatical Error Correction},
  author = {Dahlmeier, Daniel and Ng, Hwee Tou},
  booktitle = {Proceedings of the 2012 Conference of the North American Chapter of the Association for Computational Linguistics: Human Language Technologies},
  pages = {568--572},
  year = {2012},
  address = {Montr{\'e}al, Canada},
  publisher = {Association for Computational Linguistics},
  url = {https://aclanthology.org/N12-1067/}
}

@inproceedings{malmi-etal-2019-encode,
  title = {Encode, Tag, Realize: High-Precision Text Editing},
  author = {Malmi, Eric and Krause, Sebastian and Rothe, Sascha and Mirylenka, Daniil and Severyn, Aliaksei},
  booktitle = {Proceedings of the 2019 Conference on Empirical Methods in Natural Language Processing and the 9th International Joint Conference on Natural Language Processing},
  pages = {5054--5065},
  year = {2019},
  address = {Hong Kong, China},
  publisher = {Association for Computational Linguistics},
  doi = {10.18653/v1/D19-1510},
  url = {https://aclanthology.org/D19-1510/}
}

@inproceedings{awasthi-etal-2019-parallel,
  title = {Parallel Iterative Edit Models for Local Sequence Transduction},
  author = {Awasthi, Abhijeet and Sarawagi, Sunita and Goyal, Rasna and Ghosh, Sabyasachi and Piratla, Vihari},
  booktitle = {Proceedings of the 2019 Conference on Empirical Methods in Natural Language Processing and the 9th International Joint Conference on Natural Language Processing},
  pages = {4260--4270},
  year = {2019},
  address = {Hong Kong, China},
  publisher = {Association for Computational Linguistics},
  doi = {10.18653/v1/D19-1435},
  url = {https://aclanthology.org/D19-1435/}
}

@inproceedings{stahlberg-kumar-2020-seq2edits,
  title = {{Seq2Edits}: Sequence Transduction Using Span-level Edit Operations},
  author = {Stahlberg, Felix and Kumar, Shankar},
  booktitle = {Proceedings of the 2020 Conference on Empirical Methods in Natural Language Processing},
  pages = {5147--5159},
  year = {2020},
  address = {Online},
  publisher = {Association for Computational Linguistics},
  doi = {10.18653/v1/2020.emnlp-main.418},
  url = {https://aclanthology.org/2020.emnlp-main.418/}
}

@inproceedings{omelianchuk-etal-2020-gector,
  title = {{GECToR}: Grammatical Error Correction: Tag, Not Rewrite},
  author = {Omelianchuk, Kostiantyn and Atrasevych, Vitaliy and Chernodub, Artem and Skurzhanskyi, Oleksandr},
  booktitle = {Proceedings of the Fifteenth Workshop on Innovative Use of NLP for Building Educational Applications},
  pages = {163--170},
  year = {2020},
  address = {Seattle, WA, USA},
  publisher = {Association for Computational Linguistics},
  doi = {10.18653/v1/2020.bea-1.16},
  url = {https://aclanthology.org/2020.bea-1.16/}
}

@inproceedings{mohit-etal-2014-first,
  title = {The First {QALB} Shared Task on Automatic Text Correction for {Arabic}},
  author = {Mohit, Behrang and Rozovskaya, Alla and Habash, Nizar and Zaghouani, Wajdi and Obeid, Ossama},
  booktitle = {Proceedings of the EMNLP 2014 Workshop on Arabic Natural Language Processing},
  pages = {39--47},
  year = {2014},
  address = {Doha, Qatar},
  publisher = {Association for Computational Linguistics},
  doi = {10.3115/v1/W14-3605},
  url = {https://aclanthology.org/W14-3605/}
}

@inproceedings{rozovskaya-etal-2015-second,
  title = {The Second {QALB} Shared Task on Automatic Text Correction for {Arabic}},
  author = {Rozovskaya, Alla and Bouamor, Houda and Habash, Nizar and Zaghouani, Wajdi and Obeid, Ossama and Mohit, Behrang},
  booktitle = {Proceedings of the Second Workshop on Arabic Natural Language Processing},
  pages = {26--35},
  year = {2015},
  address = {Beijing, China},
  publisher = {Association for Computational Linguistics},
  doi = {10.18653/v1/W15-3204},
  url = {https://aclanthology.org/W15-3204/}
}

@inproceedings{habash-palfreyman-2022-zaebuc,
  title = {{ZAEBUC}: An Annotated {Arabic-English} Bilingual Writer Corpus},
  author = {Habash, Nizar and Palfreyman, David},
  booktitle = {Proceedings of the Thirteenth Language Resources and Evaluation Conference},
  pages = {79--88},
  year = {2022},
  address = {Marseille, France},
  publisher = {European Language Resources Association},
  url = {https://aclanthology.org/2022.lrec-1.9/}
}

@inproceedings{alhafni-etal-2023-advancements,
  title = {Advancements in {Arabic} Grammatical Error Detection and Correction: An Empirical Investigation},
  author = {Alhafni, Bashar and Inoue, Go and Khairallah, Christian and Habash, Nizar},
  booktitle = {Proceedings of the 2023 Conference on Empirical Methods in Natural Language Processing},
  pages = {6430--6448},
  year = {2023},
  address = {Singapore},
  publisher = {Association for Computational Linguistics},
  doi = {10.18653/v1/2023.emnlp-main.396},
  url = {https://aclanthology.org/2023.emnlp-main.396/}
}

@inproceedings{kwon-etal-2023-beyond,
  title = {Beyond {English}: Evaluating {LLM}s for {Arabic} Grammatical Error Correction},
  author = {Kwon, Sang and Bhatia, Gagan and Nagoudi, El Moatez Billah and Abdul-Mageed, Muhammad},
  booktitle = {Proceedings of ArabicNLP 2023},
  pages = {101--119},
  year = {2023},
  address = {Singapore},
  publisher = {Association for Computational Linguistics},
  doi = {10.18653/v1/2023.arabicnlp-1.9},
  url = {https://aclanthology.org/2023.arabicnlp-1.9/}
}

@inproceedings{bryant-etal-2017-automatic,
  title = {Automatic Annotation and Evaluation of Error Types for Grammatical Error Correction},
  author = {Bryant, Christopher and Felice, Mariano and Briscoe, Ted},
  booktitle = {Proceedings of the 55th Annual Meeting of the Association for Computational Linguistics},
  pages = {793--805},
  year = {2017},
  address = {Vancouver, Canada},
  publisher = {Association for Computational Linguistics},
  doi = {10.18653/v1/P17-1074},
  url = {https://aclanthology.org/P17-1074/}
}

@inproceedings{belkebir-habash-2021-automatic,
  title = {Automatic Error Type Annotation for {Arabic}},
  author = {Belkebir, Riadh and Habash, Nizar},
  booktitle = {Proceedings of the 25th Conference on Computational Natural Language Learning},
  pages = {596--606},
  year = {2021},
  address = {Online},
  publisher = {Association for Computational Linguistics},
  doi = {10.18653/v1/2021.conll-1.47},
  url = {https://aclanthology.org/2021.conll-1.47/}
}

@inproceedings{alhafni-habash-2025-enhancing,
  title = {Enhancing Text Editing for Grammatical Error Correction: {Arabic} as a Case Study},
  author = {Alhafni, Bashar and Habash, Nizar},
  booktitle = {Proceedings of the 63rd Annual Meeting of the Association for Computational Linguistics},
  pages = {17892--17914},
  year = {2025},
  address = {Vienna, Austria},
  publisher = {Association for Computational Linguistics},
  doi = {10.18653/v1/2025.acl-long.875},
  url = {https://aclanthology.org/2025.acl-long.875/}
}

@inproceedings{antoun-etal-2020-arabert,
  title = {{AraBERT}: Transformer-based Model for {Arabic} Language Understanding},
  author = {Antoun, Wissam and Baly, Fady and Hajj, Hazem},
  booktitle = {Proceedings of the 4th Workshop on Open-Source Arabic Corpora and Processing Tools},
  pages = {9--15},
  year = {2020},
  address = {Marseille, France},
  url = {https://aclanthology.org/2020.osact-1.2/}
}

@misc{openai-2024-gpt4o,
  title = {{GPT-4o} System Card},
  author = {{OpenAI}},
  year = {2024},
  eprint = {2410.21276},
  archivePrefix = {arXiv},
  primaryClass = {cs.CL},
  url = {https://arxiv.org/abs/2410.21276}
}

@misc{sengupta-etal-2023-jais,
  title = {{Jais} and {Jais-chat}: {Arabic}-centric Foundation and Instruction-tuned Open Generative Large Language Models},
  author = {Sengupta, Neha and Sahu, Sunil Kumar and Jia, Bokang and Katipomu, Satheesh and Li, Haonan and Koto, Fajri and Marshall, William and Gosal, Gurpreet and Liu, Cynthia and Chen, Zhimin and others},
  year = {2023},
  eprint = {2308.16149},
  archivePrefix = {arXiv},
  primaryClass = {cs.CL},
  url = {https://arxiv.org/abs/2308.16149}
}

@misc{fanar-team-2025-fanar,
  title = {Fanar: An {Arabic}-centric Multimodal Generative {AI} Platform},
  author = {{Fanar Team} and Abbas, Ummar and Ahmad, Mohammad Shahmeer and Alam, Firoj and Altinisik, Enes and Asgari, Ehsannedin and Boshmaf, Yazan and Boughorbel, Sabri and Chawla, Sanjay and Chowdhury, Shammur and others},
  year = {2025},
  eprint = {2501.13944},
  archivePrefix = {arXiv},
  primaryClass = {cs.CL},
  url = {https://arxiv.org/abs/2501.13944}
}

@inproceedings{abdul-mageed-etal-2021-arbert,
    title = "{ARBERT} {\&} {MARBERT}: Deep Bidirectional Transformers for {A}rabic",
    author = "Abdul-Mageed, Muhammad  and
      Elmadany, AbdelRahim  and
      Nagoudi, El Moatez Billah",
    editor = "Zong, Chengqing  and
      Xia, Fei  and
      Li, Wenjie  and
      Navigli, Roberto",
    booktitle = "Proceedings of the 59th Annual Meeting of the Association for Computational Linguistics and the 11th International Joint Conference on Natural Language Processing (Volume 1: Long Papers)",
    month = aug,
    year = "2021",
    address = "Online",
    publisher = "Association for Computational Linguistics",
    url = "https://aclanthology.org/2021.acl-long.551/",
    doi = "10.18653/v1/2021.acl-long.551",
    pages = "7088--7105"
}

@inproceedings{bougares-bouamor-2015-ummu,
    title = "{UMMU}@{QALB}-2015 Shared Task: Character and Word level {SMT} pipeline for Automatic Error Correction of {A}rabic Text",
    author = "Bougares, Fethi  and
      Bouamor, Houda",
    editor = "Habash, Nizar  and
      Vogel, Stephan  and
      Darwish, Kareem",
    booktitle = "Proceedings of the Second Workshop on {A}rabic Natural Language Processing",
    month = jul,
    year = "2015",
    address = "Beijing, China",
    publisher = "Association for Computational Linguistics",
    url = "https://aclanthology.org/W15-3221/",
    doi = "10.18653/v1/W15-3221",
    pages = "166--172"
}

@inproceedings{dale-etal-2012-hoo,
    title = "{HOO} 2012: A Report on the Preposition and Determiner Error Correction Shared Task",
    author = "Dale, Robert  and
      Anisimoff, Ilya  and
      Narroway, George",
    editor = "Tetreault, Joel  and
      Burstein, Jill  and
      Leacock, Claudia",
    booktitle = "Proceedings of the Seventh Workshop on Building Educational Applications Using {NLP}",
    month = jun,
    year = "2012",
    address = "Montr{\'e}al, Canada",
    publisher = "Association for Computational Linguistics",
    url = "https://aclanthology.org/W12-2006/",
    pages = "54--62"
}

@inproceedings{davis-etal-2024-prompting,
    title = "Prompting open-source and commercial language models for grammatical error correction of {E}nglish learner text",
    author = "Davis, Christopher  and
      Caines, Andrew  and
      Andersen, {\O}istein E.  and
      Taslimipoor, Shiva  and
      Yannakoudakis, Helen  and
      Yuan, Zheng  and
      Bryant, Christopher  and
      Rei, Marek  and
      Buttery, Paula",
    editor = "Ku, Lun-Wei  and
      Martins, Andre  and
      Srikumar, Vivek",
    booktitle = "Findings of the Association for Computational Linguistics: ACL 2024",
    month = aug,
    year = "2024",
    address = "Bangkok, Thailand",
    publisher = "Association for Computational Linguistics",
    url = "https://aclanthology.org/2024.findings-acl.711/",
    doi = "10.18653/v1/2024.findings-acl.711",
    pages = "11952--11967"
}

@inproceedings{eskander-etal-2013-processing,
    title = "Processing Spontaneous Orthography",
    author = "Eskander, Ramy  and
      Habash, Nizar  and
      Rambow, Owen  and
      Tomeh, Nadi",
    editor = "Vanderwende, Lucy  and
      Daum{\'e} III, Hal  and
      Kirchhoff, Katrin",
    booktitle = "Proceedings of the 2013 Conference of the North {A}merican Chapter of the Association for Computational Linguistics: Human Language Technologies",
    month = jun,
    year = "2013",
    address = "Atlanta, Georgia",
    publisher = "Association for Computational Linguistics",
    url = "https://aclanthology.org/N13-1066/",
    pages = "585--595"
}

@inproceedings{farra-etal-2014-generalized,
    title = "Generalized Character-Level Spelling Error Correction",
    author = "Farra, Noura  and
      Tomeh, Nadi  and
      Rozovskaya, Alla  and
      Habash, Nizar",
    editor = "Toutanova, Kristina  and
      Wu, Hua",
    booktitle = "Proceedings of the 52nd Annual Meeting of the Association for Computational Linguistics (Volume 2: Short Papers)",
    month = jun,
    year = "2014",
    address = "Baltimore, Maryland",
    publisher = "Association for Computational Linguistics",
    url = "https://aclanthology.org/P14-2027/",
    doi = "10.3115/v1/P14-2027",
    pages = "161--167"
}

@inproceedings{felice-etal-2016-automatic,
    title = "Automatic Extraction of Learner Errors in {ESL} Sentences Using Linguistically Enhanced Alignments",
    author = "Felice, Mariano  and
      Bryant, Christopher  and
      Briscoe, Ted",
    editor = "Matsumoto, Yuji  and
      Prasad, Rashmi",
    booktitle = "Proceedings of {COLING} 2016, the 26th International Conference on Computational Linguistics: Technical Papers",
    month = dec,
    year = "2016",
    address = "Osaka, Japan",
    publisher = "The COLING 2016 Organizing Committee",
    url = "https://aclanthology.org/C16-1079/",
    pages = "825--835"
}

@inproceedings{grundkiewicz-etal-2019-neural,
    title = "Neural Grammatical Error Correction Systems with Unsupervised Pre-training on Synthetic Data",
    author = "Grundkiewicz, Roman  and
      Junczys-Dowmunt, Marcin  and
      Heafield, Kenneth",
    editor = "Yannakoudakis, Helen  and
      Kochmar, Ekaterina  and
      Leacock, Claudia  and
      Madnani, Nitin  and
      Pil{\'a}n, Ildik{\'o}  and
      Zesch, Torsten",
    booktitle = "Proceedings of the Fourteenth Workshop on Innovative Use of NLP for Building Educational Applications",
    month = aug,
    year = "2019",
    address = "Florence, Italy",
    publisher = "Association for Computational Linguistics",
    url = "https://aclanthology.org/W19-4427/",
    doi = "10.18653/v1/W19-4427",
    pages = "252--263"
}

@inproceedings{inoue-etal-2021-interplay,
    title = "The Interplay of Variant, Size, and Task Type in {A}rabic Pre-trained Language Models",
    author = "Inoue, Go  and
      Alhafni, Bashar  and
      Baimukan, Nurpeiis  and
      Bouamor, Houda  and
      Habash, Nizar",
    editor = "Habash, Nizar  and
      Bouamor, Houda  and
      Hajj, Hazem  and
      Magdy, Walid  and
      Zaghouani, Wajdi  and
      Bougares, Fethi  and
      Tomeh, Nadi  and
      Abu Farha, Ibrahim  and
      Touileb, Samia",
    booktitle = "Proceedings of the Sixth Arabic Natural Language Processing Workshop",
    month = apr,
    year = "2021",
    address = "Kyiv, Ukraine (Virtual)",
    publisher = "Association for Computational Linguistics",
    url = "https://aclanthology.org/2021.wanlp-1.10/",
    pages = "92--104"
}

@inproceedings{jeblee-etal-2014-cmuq,
    title = "{CMUQ}@{QALB}-2014: An {SMT}-based System for Automatic {A}rabic Error Correction",
    author = "Jeblee, Serena  and
      Bouamor, Houda  and
      Zaghouani, Wajdi  and
      Oflazer, Kemal",
    editor = "Habash, Nizar  and
      Vogel, Stephan",
    booktitle = "Proceedings of the {EMNLP} 2014 Workshop on {A}rabic Natural Language Processing ({ANLP})",
    month = oct,
    year = "2014",
    address = "Doha, Qatar",
    publisher = "Association for Computational Linguistics",
    url = "https://aclanthology.org/W14-3618/",
    doi = "10.3115/v1/W14-3618",
    pages = "137--142"
}

@inproceedings{junczys-dowmunt-etal-2018-approaching,
    title = "Approaching Neural Grammatical Error Correction as a Low-Resource Machine Translation Task",
    author = "Junczys-Dowmunt, Marcin  and
      Grundkiewicz, Roman  and
      Guha, Shubha  and
      Heafield, Kenneth",
    editor = "Walker, Marilyn  and
      Ji, Heng  and
      Stent, Amanda",
    booktitle = "Proceedings of the 2018 Conference of the North {A}merican Chapter of the Association for Computational Linguistics: Human Language Technologies, Volume 1 (Long Papers)",
    month = jun,
    year = "2018",
    address = "New Orleans, Louisiana",
    publisher = "Association for Computational Linguistics",
    url = "https://aclanthology.org/N18-1055/",
    doi = "10.18653/v1/N18-1055",
    pages = "595--606"
}

@inproceedings{junczys-dowmunt-grundkiewicz-2016-phrase,
    title = "Phrase-based Machine Translation is State-of-the-Art for Automatic Grammatical Error Correction",
    author = "Junczys-Dowmunt, Marcin  and
      Grundkiewicz, Roman",
    editor = "Su, Jian  and
      Duh, Kevin  and
      Carreras, Xavier",
    booktitle = "Proceedings of the 2016 Conference on Empirical Methods in Natural Language Processing",
    month = nov,
    year = "2016",
    address = "Austin, Texas",
    publisher = "Association for Computational Linguistics",
    url = "https://aclanthology.org/D16-1161/",
    doi = "10.18653/v1/D16-1161",
    pages = "1546--1556"
}

@inproceedings{kaneko-etal-2020-encoder,
    title = "Encoder-Decoder Models Can Benefit from Pre-trained Masked Language Models in Grammatical Error Correction",
    author = "Kaneko, Masahiro  and
      Mita, Masato  and
      Kiyono, Shun  and
      Suzuki, Jun  and
      Inui, Kentaro",
    editor = "Jurafsky, Dan  and
      Chai, Joyce  and
      Schluter, Natalie  and
      Tetreault, Joel",
    booktitle = "Proceedings of the 58th Annual Meeting of the Association for Computational Linguistics",
    month = jul,
    year = "2020",
    address = "Online",
    publisher = "Association for Computational Linguistics",
    url = "https://aclanthology.org/2020.acl-main.391/",
    doi = "10.18653/v1/2020.acl-main.391",
    pages = "4248--4254"
}

@inproceedings{katsumata-komachi-2020-stronger,
    title = "Stronger Baselines for Grammatical Error Correction Using a Pretrained Encoder-Decoder Model",
    author = "Katsumata, Satoru  and
      Komachi, Mamoru",
    editor = "Wong, Kam-Fai  and
      Knight, Kevin  and
      Wu, Hua",
    booktitle = "Proceedings of the 1st Conference of the Asia-Pacific Chapter of the Association for Computational Linguistics and the 10th International Joint Conference on Natural Language Processing",
    month = dec,
    year = "2020",
    address = "Suzhou, China",
    publisher = "Association for Computational Linguistics",
    url = "https://aclanthology.org/2020.aacl-main.83/",
    pages = "827--832"
}

@inproceedings{kiyono-etal-2019-empirical,
    title = "An Empirical Study of Incorporating Pseudo Data into Grammatical Error Correction",
    author = "Kiyono, Shun  and
      Suzuki, Jun  and
      Mita, Masato  and
      Mizumoto, Tomoya  and
      Inui, Kentaro",
    editor = "Inui, Kentaro  and
      Jiang, Jing  and
      Ng, Vincent  and
      Wan, Xiaojun",
    booktitle = "Proceedings of the 2019 Conference on Empirical Methods in Natural Language Processing and the 9th International Joint Conference on Natural Language Processing (EMNLP-IJCNLP)",
    month = nov,
    year = "2019",
    address = "Hong Kong, China",
    publisher = "Association for Computational Linguistics",
    url = "https://aclanthology.org/D19-1119/",
    doi = "10.18653/v1/D19-1119",
    pages = "1236--1242"
}

@inproceedings{loem-etal-2023-exploring,
    title = "Exploring Effectiveness of {GPT}-3 in Grammatical Error Correction: A Study on Performance and Controllability in Prompt-Based Methods",
    author = "Loem, Mengsay  and
      Kaneko, Masahiro  and
      Takase, Sho  and
      Okazaki, Naoaki",
    editor = "Kochmar, Ekaterina  and
      Bexte, Marie  and
      Burstein, Jill  and
      Horbach, Andrea  and
      Laarmann-Quante, Ronja  and
      Madnani, Nitin  and
      Tack, Ana{\"\i}s  and
      Yaneva, Victoria  and
      Yuan, Zheng  and
      Zesch, Torsten",
    booktitle = "Proceedings of the 18th Workshop on Innovative Use of NLP for Building Educational Applications (BEA 2023)",
    month = jul,
    year = "2023",
    address = "Toronto, Canada",
    publisher = "Association for Computational Linguistics",
    url = "https://aclanthology.org/2023.bea-1.18/",
    doi = "10.18653/v1/2023.bea-1.18",
    pages = "205--219"
}

@inproceedings{mallinson-etal-2020-felix,
    title = "{FELIX}: Flexible Text Editing Through Tagging and Insertion",
    author = "Mallinson, Jonathan  and
      Severyn, Aliaksei  and
      Malmi, Eric  and
      Garrido, Guillermo",
    editor = "Webber, Bonnie  and
      Cohn, Trevor  and
      He, Yulan  and
      Liu, Yang",
    booktitle = "Findings of the Association for Computational Linguistics: EMNLP 2020",
    month = nov,
    year = "2020",
    address = "Online",
    publisher = "Association for Computational Linguistics",
    url = "https://aclanthology.org/2020.findings-emnlp.111/",
    doi = "10.18653/v1/2020.findings-emnlp.111",
    pages = "1244--1255"
}

@inproceedings{mallinson-etal-2022-edit5,
    title = "{E}di{T}5: Semi-Autoregressive Text Editing with {T}5 Warm-Start",
    author = "Mallinson, Jonathan  and
      Adamek, Jakub  and
      Malmi, Eric  and
      Severyn, Aliaksei",
    editor = "Goldberg, Yoav  and
      Kozareva, Zornitsa  and
      Zhang, Yue",
    booktitle = "Findings of the Association for Computational Linguistics: EMNLP 2022",
    month = dec,
    year = "2022",
    address = "Abu Dhabi, United Arab Emirates",
    publisher = "Association for Computational Linguistics",
    url = "https://aclanthology.org/2022.findings-emnlp.156/",
    doi = "10.18653/v1/2022.findings-emnlp.156",
    pages = "2126--2138"
}

@inproceedings{mesham-etal-2023-extended,
    title = "An Extended Sequence Tagging Vocabulary for Grammatical Error Correction",
    author = "Mesham, Stuart  and
      Bryant, Christopher  and
      Rei, Marek  and
      Yuan, Zheng",
    editor = "Vlachos, Andreas  and
      Augenstein, Isabelle",
    booktitle = "Findings of the Association for Computational Linguistics: EACL 2023",
    month = may,
    year = "2023",
    address = "Dubrovnik, Croatia",
    publisher = "Association for Computational Linguistics",
    url = "https://aclanthology.org/2023.findings-eacl.119/",
    doi = "10.18653/v1/2023.findings-eacl.119",
    pages = "1608--1619"
}

@inproceedings{napoles-etal-2015-ground,
    title = "Ground Truth for Grammatical Error Correction Metrics",
    author = "Napoles, Courtney  and
      Sakaguchi, Keisuke  and
      Post, Matt  and
      Tetreault, Joel",
    editor = "Zong, Chengqing  and
      Strube, Michael",
    booktitle = "Proceedings of the 53rd Annual Meeting of the Association for Computational Linguistics and the 7th International Joint Conference on Natural Language Processing (Volume 2: Short Papers)",
    month = jul,
    year = "2015",
    address = "Beijing, China",
    publisher = "Association for Computational Linguistics",
    url = "https://aclanthology.org/P15-2097/",
    doi = "10.3115/v1/P15-2097",
    pages = "588--593"
}

@inproceedings{napoles-etal-2017-jfleg,
    title = "{JFLEG}: A Fluency Corpus and Benchmark for Grammatical Error Correction",
    author = "Napoles, Courtney  and
      Sakaguchi, Keisuke  and
      Tetreault, Joel",
    editor = "Lapata, Mirella  and
      Blunsom, Phil  and
      Koller, Alexander",
    booktitle = "Proceedings of the 15th Conference of the {E}uropean Chapter of the Association for Computational Linguistics: Volume 2, Short Papers",
    month = apr,
    year = "2017",
    address = "Valencia, Spain",
    publisher = "Association for Computational Linguistics",
    url = "https://aclanthology.org/E17-2037/",
    pages = "229--234"
}

@inproceedings{nawar-2015-cufe,
    title = "{CUFE}@{QALB}-2015 Shared Task: {A}rabic Error Correction System",
    author = "Nawar, Michael",
    editor = "Habash, Nizar  and
      Vogel, Stephan  and
      Darwish, Kareem",
    booktitle = "Proceedings of the Second Workshop on {A}rabic Natural Language Processing",
    month = jul,
    year = "2015",
    address = "Beijing, China",
    publisher = "Association for Computational Linguistics",
    url = "https://aclanthology.org/W15-3215/",
    doi = "10.18653/v1/W15-3215",
    pages = "133--137"
}

@inproceedings{ng-etal-2013-conll,
    title = "The {C}o{NLL}-2013 Shared Task on Grammatical Error Correction",
    author = "Ng, Hwee Tou  and
      Wu, Siew Mei  and
      Wu, Yuanbin  and
      Hadiwinoto, Christian  and
      Tetreault, Joel",
    editor = "Ng, Hwee Tou  and
      Tetreault, Joel  and
      Wu, Siew Mei  and
      Wu, Yuanbin  and
      Hadiwinoto, Christian",
    booktitle = "Proceedings of the Seventeenth Conference on Computational Natural Language Learning: Shared Task",
    month = aug,
    year = "2013",
    address = "Sofia, Bulgaria",
    publisher = "Association for Computational Linguistics",
    url = "https://aclanthology.org/W13-3601/",
    pages = "1--12"
}

@inproceedings{obeid-etal-2020-camel,
    title = "{CAM}e{L} Tools: An Open Source Python Toolkit for {A}rabic Natural Language Processing",
    author = "Obeid, Ossama  and
      Zalmout, Nasser  and
      Khalifa, Salam  and
      Taji, Dima  and
      Oudah, Mai  and
      Alhafni, Bashar  and
      Inoue, Go  and
      Eryani, Fadhl  and
      Erdmann, Alexander  and
      Habash, Nizar",
    editor = "Calzolari, Nicoletta  and
      B{\'e}chet, Fr{\'e}d{\'e}ric  and
      Blache, Philippe  and
      Choukri, Khalid  and
      Cieri, Christopher  and
      Declerck, Thierry  and
      Goggi, Sara  and
      Isahara, Hitoshi  and
      Maegaard, Bente  and
      Mariani, Joseph  and
      Mazo, H{\'e}l{\`e}ne  and
      Moreno, Asuncion  and
      Odijk, Jan  and
      Piperidis, Stelios",
    booktitle = "Proceedings of the Twelfth Language Resources and Evaluation Conference",
    month = may,
    year = "2020",
    address = "Marseille, France",
    publisher = "European Language Resources Association",
    url = "https://aclanthology.org/2020.lrec-1.868/",
    pages = "7022--7032",
    language = "eng",
    ISBN = "979-10-95546-34-4"
}

@inproceedings{obeid-etal-2022-camelira,
    title = "{C}amelira: An {A}rabic Multi-Dialect Morphological Disambiguator",
    author = "Obeid, Ossama  and
      Inoue, Go  and
      Habash, Nizar",
    editor = "He, Yulan  and
      Ji, Heng  and
      Li, Sujian  and
      Liu, Yang  and
      Chang, Chua-Hui",
    booktitle = "Proceedings of the 2022 Conference on Empirical Methods in Natural Language Processing: System Demonstrations",
    month = dec,
    year = "2022",
    address = "Abu Dhabi, UAE",
    publisher = "Association for Computational Linguistics",
    url = "https://aclanthology.org/2022.emnlp-demos.32/",
    doi = "10.18653/v1/2022.emnlp-demos.32",
    pages = "319--326"
}

@inproceedings{raheja-etal-2023-coedit,
    title = "{C}o{E}di{T}: Text Editing by Task-Specific Instruction Tuning",
    author = "Raheja, Vipul  and
      Kumar, Dhruv  and
      Koo, Ryan  and
      Kang, Dongyeop",
    editor = "Bouamor, Houda  and
      Pino, Juan  and
      Bali, Kalika",
    booktitle = "Findings of the Association for Computational Linguistics: EMNLP 2023",
    month = dec,
    year = "2023",
    address = "Singapore",
    publisher = "Association for Computational Linguistics",
    url = "https://aclanthology.org/2023.findings-emnlp.350/",
    doi = "10.18653/v1/2023.findings-emnlp.350",
    pages = "5274--5291"
}

@inproceedings{raheja-etal-2024-medit,
    title = "m{E}di{T}: Multilingual Text Editing via Instruction Tuning",
    author = "Raheja, Vipul  and
      Alikaniotis, Dimitris  and
      Kulkarni, Vivek  and
      Alhafni, Bashar  and
      Kumar, Dhruv",
    editor = "Duh, Kevin  and
      Gomez, Helena  and
      Bethard, Steven",
    booktitle = "Proceedings of the 2024 Conference of the North American Chapter of the Association for Computational Linguistics: Human Language Technologies (Volume 1: Long Papers)",
    month = jun,
    year = "2024",
    address = "Mexico City, Mexico",
    publisher = "Association for Computational Linguistics",
    url = "https://aclanthology.org/2024.naacl-long.56/",
    doi = "10.18653/v1/2024.naacl-long.56",
    pages = "979--1001"
}

@inproceedings{rothe-etal-2021-simple,
    title = "A Simple Recipe for Multilingual Grammatical Error Correction",
    author = "Rothe, Sascha  and
      Mallinson, Jonathan  and
      Malmi, Eric  and
      Krause, Sebastian  and
      Severyn, Aliaksei",
    editor = "Zong, Chengqing  and
      Xia, Fei  and
      Li, Wenjie  and
      Navigli, Roberto",
    booktitle = "Proceedings of the 59th Annual Meeting of the Association for Computational Linguistics and the 11th International Joint Conference on Natural Language Processing (Volume 2: Short Papers)",
    month = aug,
    year = "2021",
    address = "Online",
    publisher = "Association for Computational Linguistics",
    url = "https://aclanthology.org/2021.acl-short.89/",
    doi = "10.18653/v1/2021.acl-short.89",
    pages = "702--707"
}

@inproceedings{rozovskaya-etal-2014-columbia,
    title = "The {C}olumbia System in the {QALB}-2014 Shared Task on {A}rabic Error Correction",
    author = "Rozovskaya, Alla  and
      Habash, Nizar  and
      Eskander, Ramy  and
      Farra, Noura  and
      Salloum, Wael",
    editor = "Habash, Nizar  and
      Vogel, Stephan",
    booktitle = "Proceedings of the {EMNLP} 2014 Workshop on {A}rabic Natural Language Processing ({ANLP})",
    month = oct,
    year = "2014",
    address = "Doha, Qatar",
    publisher = "Association for Computational Linguistics",
    url = "https://aclanthology.org/W14-3622/",
    doi = "10.3115/v1/W14-3622",
    pages = "160--164"
}

@inproceedings{straka-etal-2021-character,
    title = "Character Transformations for Non-Autoregressive {GEC} Tagging",
    author = "Straka, Milan  and
      N{\'a}plava, Jakub  and
      Strakov{\'a}, Jana",
    editor = "Xu, Wei  and
      Ritter, Alan  and
      Baldwin, Tim  and
      Rahimi, Afshin",
    booktitle = "Proceedings of the Seventh Workshop on Noisy User-generated Text (W-NUT 2021)",
    month = nov,
    year = "2021",
    address = "Online",
    publisher = "Association for Computational Linguistics",
    url = "https://aclanthology.org/2021.wnut-1.46/",
    doi = "10.18653/v1/2021.wnut-1.46",
    pages = "417--422"
}

@inproceedings{yuan-briscoe-2016-grammatical,
    title = "Grammatical error correction using neural machine translation",
    author = "Yuan, Zheng  and
      Briscoe, Ted",
    editor = "Knight, Kevin  and
      Nenkova, Ani  and
      Rambow, Owen",
    booktitle = "Proceedings of the 2016 Conference of the North {A}merican Chapter of the Association for Computational Linguistics: Human Language Technologies",
    month = jun,
    year = "2016",
    address = "San Diego, California",
    publisher = "Association for Computational Linguistics",
    url = "https://aclanthology.org/N16-1042/",
    doi = "10.18653/v1/N16-1042",
    pages = "380--386"
}

@inproceedings{yuan-etal-2021-multi,
    title = "Multi-Class Grammatical Error Detection for Correction: A Tale of Two Systems",
    author = "Yuan, Zheng  and
      Taslimipoor, Shiva  and
      Davis, Christopher  and
      Bryant, Christopher",
    editor = "Moens, Marie-Francine  and
      Huang, Xuanjing  and
      Specia, Lucia  and
      Yih, Scott Wen-tau",
    booktitle = "Proceedings of the 2021 Conference on Empirical Methods in Natural Language Processing",
    month = nov,
    year = "2021",
    address = "Online and Punta Cana, Dominican Republic",
    publisher = "Association for Computational Linguistics",
    url = "https://aclanthology.org/2021.emnlp-main.687/",
    doi = "10.18653/v1/2021.emnlp-main.687",
    pages = "8722--8736"
}

@inproceedings{zaghouani-etal-2014-large,
    title = "Large Scale {A}rabic Error Annotation: Guidelines and Framework",
    author = "Zaghouani, Wajdi  and
      Mohit, Behrang  and
      Habash, Nizar  and
      Obeid, Ossama  and
      Tomeh, Nadi  and
      Rozovskaya, Alla  and
      Farra, Noura  and
      Alkuhlani, Sarah  and
      Oflazer, Kemal",
    editor = "Calzolari, Nicoletta  and
      Choukri, Khalid  and
      Declerck, Thierry  and
      Loftsson, Hrafn  and
      Maegaard, Bente  and
      Mariani, Joseph  and
      Moreno, Asuncion  and
      Odijk, Jan  and
      Piperidis, Stelios",
    booktitle = "Proceedings of the Ninth International Conference on Language Resources and Evaluation ({LREC}'14)",
    month = may,
    year = "2014",
    address = "Reykjavik, Iceland",
    publisher = "European Language Resources Association (ELRA)",
    url = "https://aclanthology.org/L14-1721/"
}

@inproceedings{zaghouani-etal-2015-correction,
    title = "Correction Annotation for Non-Native {A}rabic Texts: Guidelines and Corpus",
    author = "Zaghouani, Wajdi  and
      Habash, Nizar  and
      Bouamor, Houda  and
      Rozovskaya, Alla  and
      Mohit, Behrang  and
      Heider, Abeer  and
      Oflazer, Kemal",
    editor = "Ide, Nancy  and
      Pustejovsky, James  and
      Goecke, Daniela  and
      Petruck, Miriam  and
      Bunt, Harry",
    booktitle = "Proceedings of the 9th Linguistic Annotation Workshop",
    month = jun,
    year = "2015",
    address = "Denver, Colorado, USA",
    publisher = "Association for Computational Linguistics",
    url = "https://aclanthology.org/W15-1614/",
    doi = "10.3115/v1/W15-1614",
    pages = "129--139"
}

@inproceedings{zhang-etal-2023-non,
    title = "Non-autoregressive Text Editing with Copy-aware Latent Alignments",
    author = "Zhang, Yu  and
      Zhang, Yue  and
      Cui, Leyang  and
      Fu, Guohong",
    editor = "Bouamor, Houda  and
      Pino, Juan  and
      Bali, Kalika",
    booktitle = "Proceedings of the 2023 Conference on Empirical Methods in Natural Language Processing",
    month = dec,
    year = "2023",
    address = "Singapore",
    publisher = "Association for Computational Linguistics",
    url = "https://aclanthology.org/2023.emnlp-main.437/",
    doi = "10.18653/v1/2023.emnlp-main.437",
    pages = "7075--7085"
}

@inproceedings{zhao-etal-2019-improving,
    title = "Improving Grammatical Error Correction via Pre-Training a Copy-Augmented Architecture with Unlabeled Data",
    author = "Zhao, Wei  and
      Wang, Liang  and
      Shen, Kewei  and
      Jia, Ruoyu  and
      Liu, Jingming",
    editor = "Burstein, Jill  and
      Doran, Christy  and
      Solorio, Thamar",
    booktitle = "Proceedings of the 2019 Conference of the North {A}merican Chapter of the Association for Computational Linguistics: Human Language Technologies, Volume 1 (Long and Short Papers)",
    month = jun,
    year = "2019",
    address = "Minneapolis, Minnesota",
    publisher = "Association for Computational Linguistics",
    url = "https://aclanthology.org/N19-1014/",
    doi = "10.18653/v1/N19-1014",
    pages = "156--165"
}

@misc{coyne-etal-2023-analyzing,
  title = {Analyzing the Performance of {GPT}-3.5 and {GPT}-4 in Grammatical Error Correction},
  author = {Coyne, Steven and Sakaguchi, Keisuke and Galvan-Sosa, Diana and Zock, Michael and Inui, Kentaro},
  year = {2023},
  eprint = {2303.14342},
  archivePrefix = {arXiv},
  primaryClass = {cs.CL},
  url = {https://arxiv.org/abs/2303.14342}
}

@misc{fang-etal-2023-chatgpt,
  title = {Is {ChatGPT} a Highly Fluent Grammatical Error Correction System? A Comprehensive Evaluation},
  author = {Fang, Tao and Yang, Shu and Lan, Kaixin and Wong, Derek F. and Hu, Jinpeng and Chao, Lidia S. and Zhang, Yue},
  year = {2023},
  eprint = {2304.01746},
  archivePrefix = {arXiv},
  primaryClass = {cs.CL},
  url = {https://arxiv.org/abs/2304.01746}
}

@misc{wu-etal-2023-chatgpt,
  title = {{ChatGPT} or {Grammarly}? Evaluating {ChatGPT} on Grammatical Error Correction Benchmark},
  author = {Wu, Haoran and Wang, Wenxuan and Wan, Yuxuan and Jiao, Wenxiang and Lyu, Michael R.},
  year = {2023},
  eprint = {2303.13648},
  archivePrefix = {arXiv},
  primaryClass = {cs.CL},
  url = {https://arxiv.org/abs/2303.13648}
}

\clearpage
\appendix

\section{Hyperparameters}
\label{sec:hyperparams}

All \textsc{STAGEET} models use AraBERTv02 as the pretrained encoder. Training uses a learning rate of $5\mathrm{e}{-5}$, batch size 32, maximum sequence length 512, seed 42, and two A100 GPUs. Each model is trained for 50 epochs. Stage-wise label pruning thresholds are set to 10, 10, 10, and 30 for spacing, orthography, morpho-lexical correction, and punctuation, respectively. The punctuation confidence threshold is selected on development data and fixed before final test evaluation.

For \textsc{STAGEET-MH}, the four stages are trained with a stage sampling ratio of 1:3:3:1. The best checkpoint is selected by development-set M2. For \mbox{\textsc{STAGEET-Sep}}, each stage is trained as a separate end-to-end tagger; during model selection, the top two candidates under M2 are retained at each stage and passed to the next stage. Models evaluated on ZAEBUC are trained on QALB-2014 plus a tenfold upsampled ZAEBUC training set.

\section{Stage-wise Data Construction}
\label{sec:stage-wise-data-construction}

\textsc{STAGEET} constructs typed supervision directly from parallel GEC data. For each raw--gold sentence pair, the raw sentence is whitespace-normalized to form the initial hypothesis $x_0$, and the gold sentence is normalized in the same way. The construction then traverses the stage inventory in order. At stage $k$, the current hypothesis $x_{k-1}$ is aligned with the gold sentence, the induced local operations are assigned to typed categories, and only the operations associated with the current stage are applied. The resulting sentence $x_k$ serves as the target for stage $k$ and as the input to the next stage.

The assignment is deterministic and ordered. Boundary-preserving whitespace edits are considered first. Remaining non-punctuation edits are then tested for local orthographic criteria. Edits that are neither boundary-only nor safely orthographic are assigned to the morpho-lexical stage, which functions as the residual non-punctuation correction stage. Punctuation operations are held out until the final stage. After each stage is applied, the sentence is realigned to the gold sentence before the next stage is constructed, so later decisions are anchored to the updated surface form rather than to tokens that may have been merged, split, deleted, or rewritten.

\begin{table*}[t]
\centering
\small
\begin{tabular}{p{0.96\linewidth}}
\toprule
\textbf{Automatic construction of typed stage supervision} \\
\midrule
\textbf{Input:} raw sentence $x$, corrected sentence $y$, stage order
$\mathcal{S}=(\mathrm{space},\mathrm{orthography},\mathrm{morpho\mbox{-}lexical},\mathrm{punctuation})$.
Normalize whitespace in $x$ and $y$ to obtain $x_0$ and $y$. \\
For each non-punctuation stage $s \in \mathcal{S}$ before punctuation, align the current hypothesis $x_{s-1}$ to $y$ and decompose the difference into atomic edit operations. Assign each operation to a stage using deterministic tests: boundary-only changes are spacing edits; local Arabic spelling and normalization changes are orthographic edits; remaining non-punctuation changes are morpho-lexical edits. Apply only the operations assigned to $s$, yielding $x_s$. If additional non-punctuation residual edits become available after earlier rewrites, assign and apply them before punctuation. \\
Before the final stage, punctuation-tokenize the current hypothesis and gold sentence. Align them again, extract punctuation insertions, deletions, and replacements, and apply only punctuation edits to obtain $x_4$. Verify that the completed staged trajectory reconstructs the gold sentence. Convert each transition $(x_{s-1},x_s)$ into subword edit-tagging examples, compress frequent edit labels, and write stage-specific training files. \\
\textbf{Output:} typed intermediate targets $x_1,\ldots,x_4$ and stage-specific edit-tagging supervision. \\
\bottomrule
\end{tabular}
\caption{Pseudocode for constructing the staged training data from ordinary raw--corrected sentence pairs.}
\label{tab:stage-construction-pseudocode}
\end{table*}

The spacing stage is applied first because token-boundary errors determine the units available to later stages. It identifies boundary-only transformations between the current hypothesis and the gold sentence, namely transformations in which the sequence of non-space characters is unchanged. These transformations cover split, merge, and whitespace insertion or deletion patterns. Once a valid boundary candidate is identified, the corresponding atomic operations are applied before lexical or punctuation edits are considered.

The orthography stage applies local spelling and normalization edits. Edits are assigned to this stage when they match Arabic orthographic confusions, hamza-seat variants, alif-related alternations, adjacent transpositions, common spelling replacements, or short single-edit spelling changes after normalization. The stage is restricted to local form changes so that broader lexical or morpho-syntactic replacements are not absorbed into the orthographic label space.

The morpho-lexical stage receives non-punctuation edits that are not safely categorized as spacing or orthography. It includes edits involving proclitics, agreement suffixes, and other lexical substitutions, insertions, and deletions. To keep the staged reconstruction complete, remaining non-punctuation edits that become classifiable after earlier stages are assigned before punctuation is introduced. This handles cases in which a local operation becomes unambiguous only after preceding boundary or orthographic changes have been applied.

Punctuation is applied last. Operations whose source or target consists of punctuation are assigned to this stage, covering punctuation insertion, deletion, and replacement. Before punctuation supervision is extracted, the intermediate sentence is punctuation-tokenized. This keeps punctuation edits separate from neighboring lexical tokens and matches the punctuation-aware form used during scoring. The final staged output is verified against the gold sentence under the same tokenization.

After the intermediate sentence pairs are built, each stage is converted into subword edit-tagging examples using the same edit-label representation as the models. Labels are also written in a compressed form for training. Rare labels are pruned during model training, with thresholds 10, 10, 10, and 30 for spacing, orthography, morpho-lexical correction, and punctuation, respectively. Table~\ref{tab:constructed-data-stats} summarizes the constructed compressed training files used in the experiments.

\begin{table*}[t]
\centering
\small
\setlength{\tabcolsep}{4pt}
\begin{tabular}{llrrrr}
\toprule
Dataset & Split & Space & Orth. & Morph. & Punc. \\
\midrule
QALB-2014 & Train & 1.26M / 72 & 1.24M / 1339 & 1.14M / 2621 & 1.19M / 150 \\
ZAEBUC & Train & 28.4K / 11 & 27.8K / 230 & 20.6K / 203 & 24.7K / 26 \\
\bottomrule
\end{tabular}
\caption{Constructed compressed stage data. Each cell reports the number of subword training instances and compressed label types before training-time pruning.}
\label{tab:constructed-data-stats}
\end{table*}

\section{Stage Label Examples}
\label{sec:stage-label-examples}

Table~\ref{tab:stage-label-examples} lists the five most frequent compressed labels in each QALB-2014 training stage. The distributions are dominated by \texttt{K*}, but the most frequent non-KEEP labels differ substantially across stages.

\begin{table}[t]
\centering
\small
\setlength{\tabcolsep}{4pt}
\begin{tabular}{llrr}
\toprule
Stage & Label & Count & Freq. \\
\midrule
Spacing & \texttt{K*} & 1,208,308 & 97.60 \\
Spacing & \texttt{MK*} & 17,620 & 1.42 \\
Spacing & \texttt{I\_[ ]K*} & 5,943 & 0.48 \\
Spacing & \texttt{KKI\_[ ]K*} & 4,490 & 0.36 \\
Spacing & \texttt{KKKI\_[ ]K*} & 712 & 0.06 \\
\midrule
Orthography & \texttt{K*} & 1,073,059 & 87.68 \\
Orthography & \texttt{R\_[}\arpayload{'a}\texttt{]K*} & 40,556 & 3.31 \\
Orthography & \texttt{R\_[}\arpayload{'i}\texttt{]K*} & 16,404 & 1.34 \\
Orthography & \texttt{KKR\_[}\arpayload{'a}\texttt{]K*} & 10,370 & 0.85 \\
Orthography & \texttt{R\_[}\arpayload{T}\texttt{]} & 10,053 & 0.82 \\
\midrule
Morpho-lexical & \texttt{K*} & 1,095,213 & 97.53 \\
Morpho-lexical & \texttt{D*} & 3,254 & 0.29 \\
Morpho-lexical & \texttt{K*I\_[}\arpayload{A}\texttt{]} & 3,154 & 0.28 \\
Morpho-lexical & \texttt{MK*} & 1,546 & 0.14 \\
Morpho-lexical & \texttt{DK*} & 1,454 & 0.13 \\
\midrule
Punctuation & \texttt{K*} & 1,054,323 & 90.20 \\
Punctuation & \texttt{K*A\_[}\arpayload{,}\texttt{]} & 62,207 & 5.32 \\
Punctuation & \texttt{K*A\_[.]} & 22,970 & 1.97 \\
Punctuation & \texttt{D*} & 5,545 & 0.47 \\
Punctuation & \texttt{R\_[}\arpayload{,}\texttt{]} & 5,151 & 0.44 \\
\bottomrule
\end{tabular}
\caption{Five most frequent compressed labels in each QALB-2014 training stage. Frequencies are percentages within the corresponding stage.}
\label{tab:stage-label-examples}
\end{table}

\section{Gradient Conflict Matrices}
\label{sec:gradient-conflict-matrices}

\begin{table}[t]
\centering
\small
\setlength{\tabcolsep}{4pt}
\begin{tabular}{lrrrr}
\toprule
\multicolumn{5}{c}{Full BERT} \\
\midrule
Stage & Space & Spell. & Morph. & Punc. \\
\midrule
Space & -- & .0049 & .0059 & .0133 \\
Spell. & .0049 & -- & .0027 & .0089 \\
Morph. & .0059 & .0027 & -- & .0032 \\
Punc. & .0133 & .0089 & .0032 & -- \\
\midrule
\multicolumn{5}{c}{Last BERT layer} \\
\midrule
Stage & Space & Spell. & Morph. & Punc. \\
\midrule
Space & -- & -.0038 & -.0013 & -.0035 \\
Spell. & -.0038 & -- & -.0011 & .0011 \\
Morph. & -.0013 & -.0011 & -- & -.0022 \\
Punc. & -.0035 & .0011 & -.0022 & -- \\
\bottomrule
\end{tabular}
\caption{Pairwise average gradient cosine similarity for the shared multi-head model. Rows and columns correspond to stage losses.}
\label{tab:gradient-cosine-matrices}
\end{table}

\begin{table}[t]
\centering
\small
\setlength{\tabcolsep}{4pt}
\begin{tabular}{lrrrr}
\toprule
\multicolumn{5}{c}{Full BERT} \\
\midrule
Stage & Space & Spell. & Morph. & Punc. \\
\midrule
Space & -- & 45\% & 20\% & 20\% \\
Spell. & 45\% & -- & 45\% & 30\% \\
Morph. & 20\% & 45\% & -- & 50\% \\
Punc. & 20\% & 30\% & 50\% & -- \\
\midrule
\multicolumn{5}{c}{Last BERT layer} \\
\midrule
Stage & Space & Spell. & Morph. & Punc. \\
\midrule
Space & -- & 52\% & 50\% & 56\% \\
Spell. & 52\% & -- & 50\% & 52\% \\
Morph. & 50\% & 50\% & -- & 48\% \\
Punc. & 56\% & 52\% & 48\% & -- \\
\bottomrule
\end{tabular}
\caption{Pairwise gradient-conflict rates for the shared multi-head model. Each cell is the proportion of sampled gradients with negative cosine similarity.}
\label{tab:gradient-conflict-matrices}
\end{table}

Tables~\ref{tab:gradient-cosine-matrices} and~\ref{tab:gradient-conflict-matrices} provide the pairwise details behind the summary in Table~\ref{tab:gradient-conflict}. Average cosine similarities are close to zero throughout the encoder and become slightly negative for several stage pairs in the last BERT layer. The corresponding conflict rates indicate that negative gradient alignment is not concentrated in a single stage pair. This pattern is consistent with the use of stage-specific adapters rather than routing all stage losses through a fully shared encoder representation.

\section{LLM Prompt}
\label{sec:llm-prompt}

\begin{figure*}[!t]
\centering
\includegraphics[width=\textwidth]{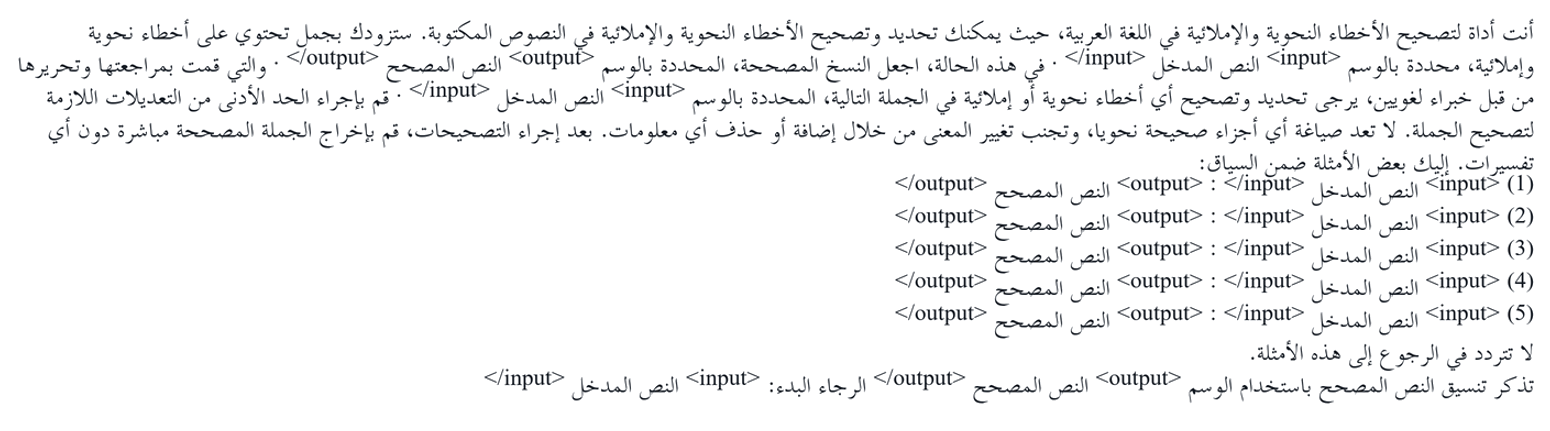}
\caption{Arabic five-shot prompt used for prompted LLM references, following SWEET \citep{alhafni-habash-2025-enhancing}.}
\label{fig:llm-prompt}
\end{figure*}

The prompted LLM references are evaluated with the Arabic five-shot prompt shown in Figure~\ref{fig:llm-prompt}, following the prompt used by SWEET \citep{alhafni-habash-2025-enhancing}. The prompt asks the model to return only the corrected sentence using the specified input and output tags.

\end{document}